\RequirePackage{fix-cm}

\documentclass{svproc}      
\usepackage{graphicx}
\usepackage{hyperref}
\usepackage[vertfit]{breakurl} 

\usepackage{url}

\begin{document}
\mainmatter  
\title{Automatic Generation of Descriptive Titles for Video Clips Using Deep Learning}

\titlerunning{Descriptive Titles for Video Clips}        

\author{Soheyla Amirian\inst{1}\inst{2} \and Khaled Rasheed\inst{1} \and Thiab R. Taha\inst{1} \and Hamid R. Arabnia\inst{1}}

\authorrunning{Soheyla Amirian et al.}

\institute{Department of Computer Science, \\The University of Georgia, Athens, GA 30602-7404, USA,\\
\and
\email{amirian@uga.edu}
}

\maketitle

\begin{abstract}
Over the last decade, the use of Deep Learning in many applications produced results that are comparable to and in some cases surpassing human expert performance. The application domains include diagnosing diseases, finance, agriculture, search engines, robot vision, and many others. In this paper, we are proposing an architecture that utilizes image/video captioning methods and Natural Language Processing systems to generate a title and a concise abstract for a video. Such a system can potentially be utilized in many application domains, including, the cinema industry, video search engines, security surveillance, video databases/warehouses, data centers, and others. The proposed system functions and operates as followed: it reads a video; representative image frames are identified and selected; the image frames are captioned; NLP is applied to all generated captions together with text summarization; and finally, a title and an abstract are generated for the video. All functions are performed automatically. Preliminary results are provided in this paper using publicly available datasets. This paper is not concerned about the efficiency of the system at the execution time. We hope to be able to address execution efficiency issues in our subsequent publications.

\end{abstract}

\section{Introduction}
\label{sec:intro}

\begin{figure}[htbp]
    \centering
    \includegraphics[width=\linewidth]{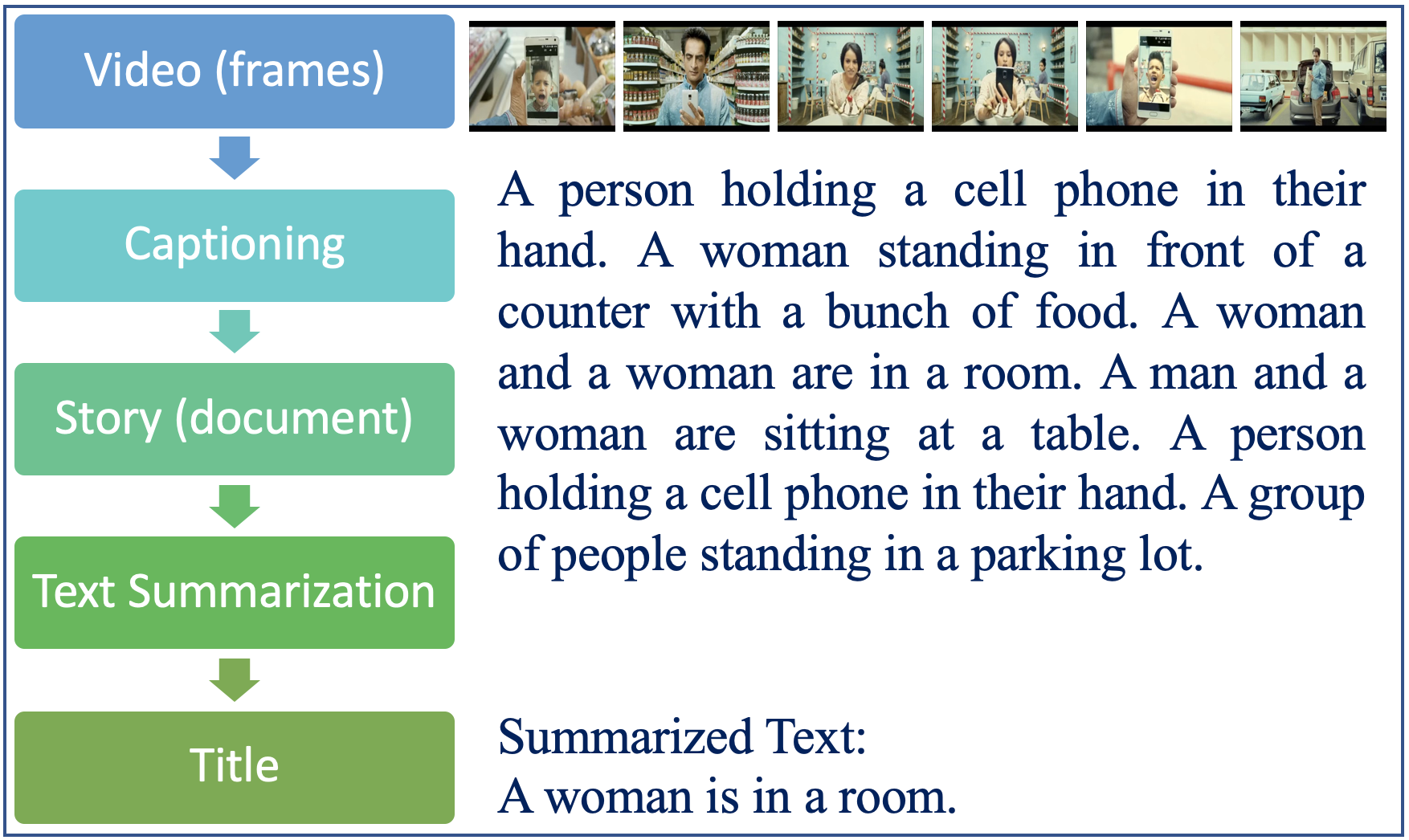}
    \caption{This is an overall example of the proposed system. The key-frames of the video are selected and captioned. The resulting document is processed with the text summarization method and the output is a possible title for the corresponding video.}
    \label{fig:method}
\end{figure}

The use of very large neural networks as Deep Learning methods that are inspired by the human brain system has recently dominated most of the researchers’ work in several domains to help in improving the results and make it more desirable for people. Machine Translation, Self-driving cars, Robotics \cite{soanssa}, Digital Marketing, Customer Services, and Better Recommendations are some applications for deep learning. In more recent years, deep learning \cite{amirian2018} has positively and significantly impacted the field of image recognition specifically, allowing much more flexibility. 
In this research, we attempt to utilize image/video captioning \cite{luo2018discriminability,sung2019adversarial} methods and Natural Language Processing systems to generate a sentence as a title for a long video that could be useful in many ways. Using an automated system instead of watching many videos to get titles could be time-saving. It can also be used in the cinema industry, search engines, and supervision cameras to name a few. We present an example of the overall process in Figure \ref{fig:method}.

Image and video captioning with deep learning are used for the difficult task of recognizing the objects and actions in an image or video and creating a succinct meaningful sentence based on the contents found. Text summarization \cite{allahyari2017text} is the task of generating a concise and fluent summary for a document(s) while preserving key information content. This paper proposes an architecture by utilizing the image/video captioning system and text summarization methods to make a title and an abstract for a long video. 
For constructing a story about a video, we extract the key-frames of the video which give more information, and then we feed those key-frames to the captioning system to make a caption or a document for them. For the captioning system, different methods as encoder-decoder or generative adversarial networks exist that propose different object detection methods. Also for the text summarization, we use both Extractive and abstractive methods \cite{paulus2017deep} to generate the title and the abstract respectively. We provide more details in the next sections.

The main contribution of this research is to explore the possibility of making a title and a concise abstract for a long video by utilizing deep learning technologies to save time through automation in many application domains. In the rest of the article, we describe the different parts of our proposed architecture which are image/video captioning and text summarization methods and provide a literature review for each component. Then, we explain the methodology of the proposed architecture and how it works. We present a proof of concept through experiments using publicly available data-sets. The article is concluded with a discussion of the results and our future work.

\section{Definition and Related Work}
\label{sec:related}

The proposed architecture in this paper consists of two main components, namely Image/Video captioning and Text summarization methods, each having different parts. In this section, we dissect each component. Also, we review some previous work about image/video captioning and text summarization that support parts of our proposed architecture.

\subsection{Image/Video Captioning:}
Describing a short video in natural language is a trivial task for most people, but a challenging one for machines. From the methodological perspective, categorizing the models or algorithms is challenging because it is difficult to assert the contributions of the visual features and the adopted language model to the final description. Automatically generating natural language sentences describing a video clip generally has two components: extracting the visual information, as Encoder; and describing it in a grammatically correct natural language sentence as Decoder. With a convolutional neural network, the objects and features are extracted from the video frames, then a neural network is used to generate a natural sentence based on the available information, on which an image captioning method would be utilized for captioning the frames \cite{soh2019image}.  

In the field of image captioning, Aneja et al. \cite{aneja2018convolutional} developed a convolutional image captioning technique with existing LSTM techniques and also analyzed the differences between RNN based learning and their method. This technique contains three main components. The first and the last components are input/output word embeddings respectively.
However, while the middle component contains LSTM or GRU units in the RNN case, masked convolutions are employed in their CNN-based approach. This component is feed-forward without any recurrent function. Their CNN with attention (Attn) achieved comparable performance. They also experimented with an attention mechanism and attention parameters using the conv-layer activations. The results of the CNN+Attn method were increased relative to the LSTM baseline. For better performance on the MSCOCO they used ResNet features and the results show that ResNet boosts their performance on the MSCOCO. The results on MSCOCO with ResNet101 and ResNet152 were impressive.

In video captioning, Krishna et al. \cite{krishna2017dense}, however, presented Dense-captioning, which focuses on detecting multiple events that occur in a video by jointly localizing temporal proposals of interest and then describing each with natural language. This model introduced a new captioning module that uses contextual information from past and future events to jointly describe all events. They implemented the model on the ActivityNet Captions dataset. The captions that came out of ActivityNet shift sentence descriptions from being object-centric in images to action-centric in videos. 
Ding et al. \cite{ding2019long} proposed novel techniques for the application of long video segmentation, which can effectively shorten the retrieval time. Redundant video frame detection based on the Spatio-temporal interest points (STIPs) and a novel super-frame segmentation are combined to improve the effectiveness of video segmentation. After that, the super-frame segmentation of the filteblue long video is performed to find an interesting clip. Keyframes from the most impactful segments are converted to video captioning by using the saliency detection and LSTM variant network. Finally, the attention mechanism is used to select more crucial information to the traditional LSTM.  Generative Adversarial Networks help to have more flexibility in these methods \cite{9071372}. Therefore, we can see that Sung Park et al. \cite{sung2019adversarial} applied Adversarial Networks in their framework. They propose to apply adversarial techniques during inference, designing a discriminator which encourages multi-sentence video description. They decouple a discriminator to evaluate visual relevance to the video, language diversity and fluency, and coherence across sentences on the ActivityNet Captions dataset.

Sequence models like recurrent neural network  (RNN)~\cite{chung2014empirical} have been widely utilized in speech recognition, natural language processing, and other areas. Sequence models can address supervised learning problems like machine translation \cite{cho2014learning}, name entity recognition, DNA sequence analysis, video activity recognition, and sentiment classification.
\textit{LSTM}, as a special RNN structure, has proven to be stable and powerful for modeling long-range dependencies in various studies. LSTM can be adopted as a building block for complex structures. The complex unit in Long Short Term Memory is called a memory cell. Each memory cell is built around a central linear unit with a fixed self-connection \cite{hochreiter1997long}. LSTM is historically proven to be more powerful and more effective than a regular RNN since it has three gates (forget, update, and output). Long Short Term Memory recurrent neural networks can be used to generate complex sequences with long-range structure \cite{kiros2014unifying,wu2016google}.

\subsection{Text Summarization:}
Automatic text summarization is the task of producing a concise and fluent summary while preserving key information content and overall meaning \cite{allahyari2017text}. Extractive and Abstractive are the two main categories of summarization algorithms.

Extractive summarization systems form summaries by copying parts of the input. Extractive summarization is implemented by identifying the important sections of the text, processing, and combining them to form a meaningful summary. 
Abstractive summarization systems generate new phrases, possibly rephrasing or using words that were not in the original text.
Abstractive summaries are generated by interpreting the raw text and generating the same information in a different and concise form by using complex neural network-based architectures such as RNNs and LSTMs.
Paulus et al. \cite{paulus2017deep} proposed a neural network model with a novel intra-attention that attends over the input and continuously generated output separately and a new training method that combines standard supervised word prediction and reinforcement learning (RL). Also, Roul et al. \cite{raffel2019exploring}
introduced the landscape of transfer learning techniques for NLP with a unified framework that converts every language problem into a text-to-text format.

Text summarization can be further divided into two categories: single and multi-text summarization. 
In single text summarization \cite{roul2019new}, the text is summarized from one document whereas Multi-document text summarization systems are able to generate reports that are rich in important information, and present varying views that span multiple documents.

\section{Methodology}
\label{sec:method}

\begin{figure*}[!ht]
    \centering
    \includegraphics[width=\linewidth]{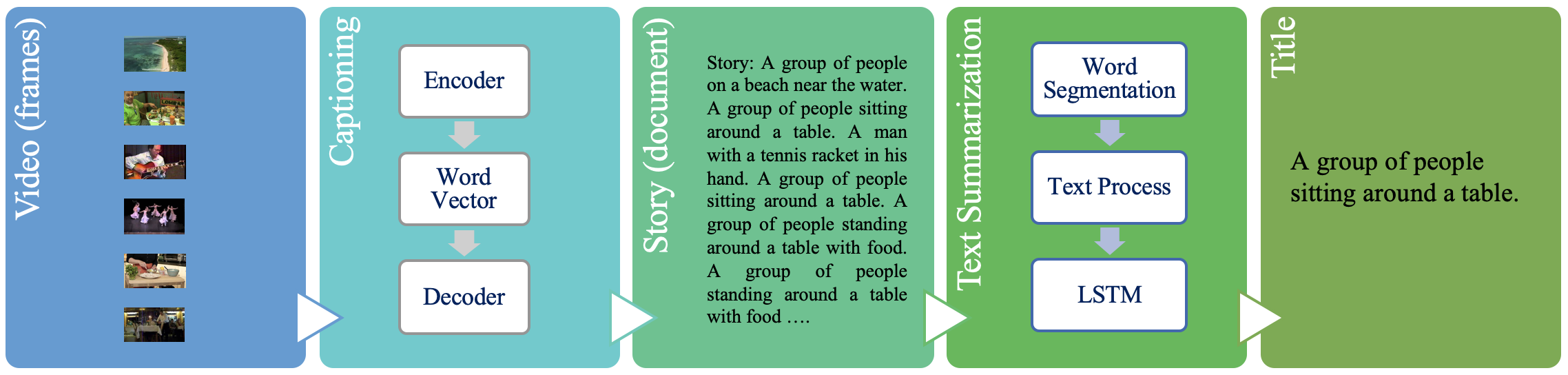}
    \caption{This is the overall architecture of the proposed method that parts to two separate process for the video captioning and text summarization.}
    \label{fig:process}
\end{figure*}

The proposed architecture consists of two different, complementary processes: Video Captioning and Text Summarization. In the first process for video captioning, the system gets a video as an input, then generates a story for the video. The generated story will feed to the second process as a document and it summarizes the document to a sentence and an abstract. Figure \ref{fig:process} shows the complete process of the suggested architecture. Further, we explain the details of each part.

\subsection{Video to Document Process:} Image/video description is the automatic generation of meaningful sentences that describes the events in an image/video (frames).
A video consists of many frames each representing an image. Some of the images/frames give much information and some are just basically repeating a scene. Therefore, we select some key-frames that include more information. The in-between frames are just repeating with subtle changes. A sequence of key-frames defines which movement the viewer will see. Therefore, the order of the key-frames on the video or animation defines the timing of the movement. 

\begin{figure}[htbp]
    \centering
    \includegraphics[width=\linewidth]{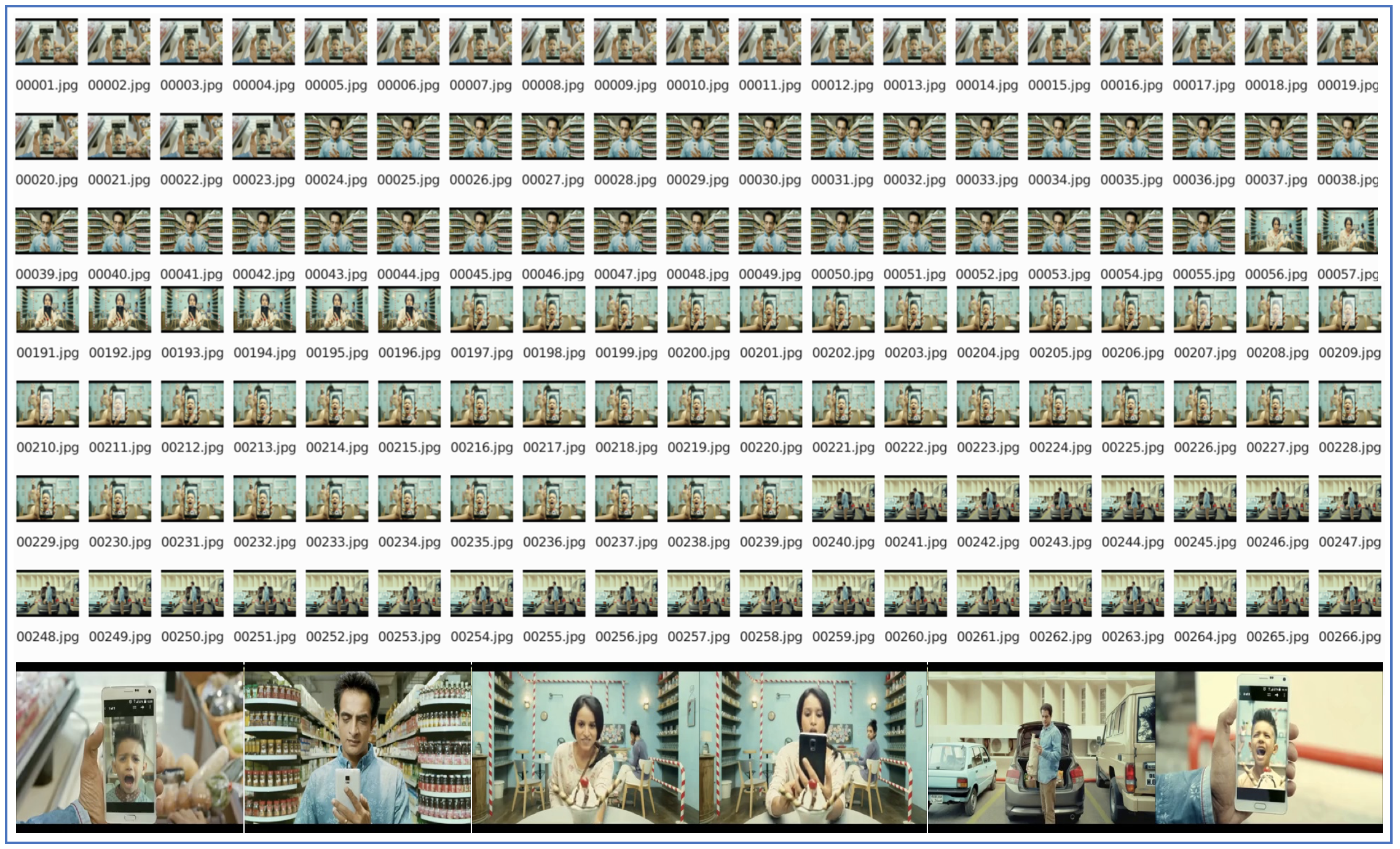}

    \caption{Video frames: in-between frames and the keyframes. We observe that many frames are repeating.}
    \label{fig:key}
\end{figure}
\begin{figure}[htbp]
    \centering
    \includegraphics[width=\linewidth]{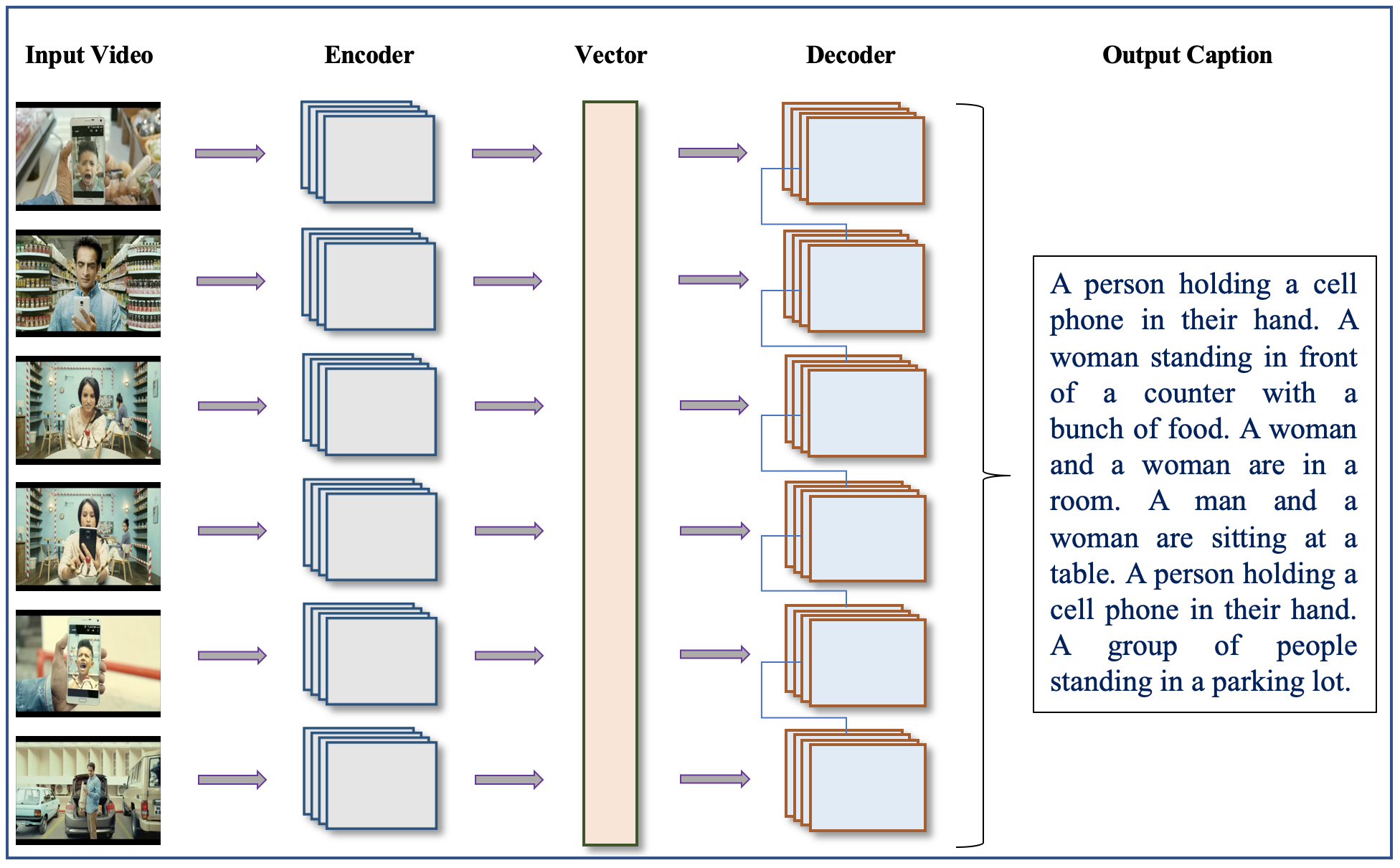}
    \caption{The task of video captioning can be divided logically into two modules: one module is based on an image-based model which extracts important features and nuances from video frames; another module is based on a language-based model, which translates the features and objects produced with the image-based model to meaningful sentences. 
}
    \label{fig:VideoC}
\end{figure}
One of our contributions in this research is doing some experiments by selecting different key-frames to have a story for long videos to see if we can have the same extracted information. So, one task is to get the key-frames and process them to be captioned, instead of using all the frames of the video to save time and resources for getting the same result. See Figure \ref{fig:key} for an illustration of the frames, key-frames, and in-between frames.

The captioning part consists of two phases: Encoder and Decoder. 
The Encoder part extracts the image information using convolutional neural networks like object detection methods to extract the objects and actions and then put them in a vector. ResNet, DenseNet, RCNN series, Yolo and ... \cite{amirian2018} can be used as object detection methods. Then the vector enters the decoder phase.
The Decoder gets the vector and then with RNN methods generates a meaningful caption for the image.
These two phases could work simultaneously. Figure \ref{fig:VideoC} illustrates the captioning process.
Captions are evaluated using the BLEU, METEOR, CIDEr, and other metrics \cite{chen2015microsoft,vedantam2015cider,xu2015show}. These metrics are common for comparing the different image and video captioning models and have varying degrees of similarity with human judgment \cite{sigurdsson2016hollywood}.

\subsection{Document to Title Process:} 
\begin{figure}[htbp]
    \centering
    \includegraphics[scale=0.36]{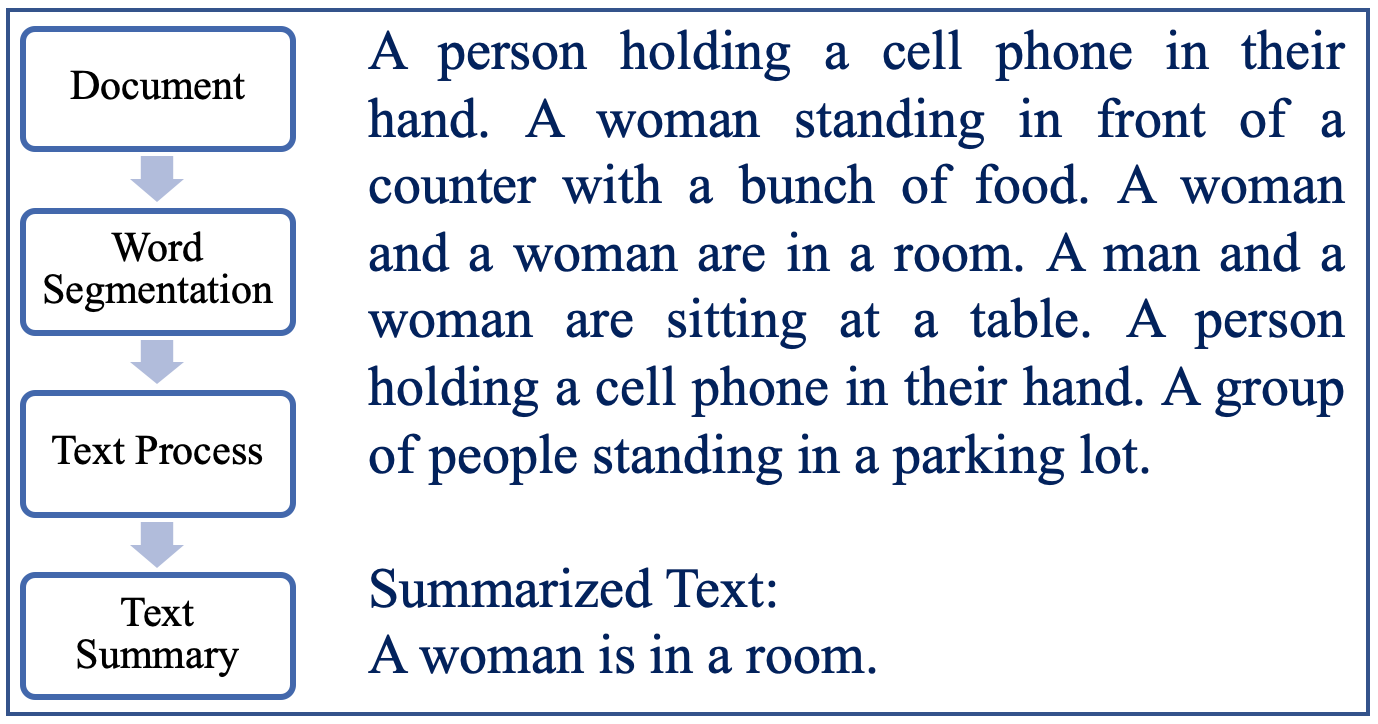}
    \caption{The output of the captioning system is a document(s) that is an input to the extractive text summarization method. Then the text would be processed to weight the words, and it shows the most likely descriptive title that the text could have.}
    \label{fig:txtS}
\end{figure}

For generating and assigning a title to the video clip, we use an extractive text summarization technique. To keep it simple, we are using an unsupervised learning approach to find the sentence similarity and rank them \cite{txt}.
The process is that we give the produced document as an input, it splits the whole document into sentences, then it removes stop words, builds a similarity matrix, generates rank based on the matrix, and at the end, it picks the top $N$ sentences for a descriptive title. Figure \ref{fig:txtS} shows an example of the extractive text summarization system.
Also for having an abstract, we implemented and used the abstractive text summarization method for the video \cite{absTxt}. Abstractive summarization methods interpret and examine the text by using advanced natural language techniques in order to generate a new shorter text that conveys the most important information from the original text \cite{allahyari2017text}.

\section{Experiments}
\label{sec:research}
The main goal of our experiments is to evaluate the utility of the proposed architecture as a proof of concept.
For implementing our idea, first, we need to get a story as a document from the image/video captioning model. So, for a given video we explore the frames, then feed the selected key-frames to the system to get the description. Implementing this part, captions have been generated by the by Luo et al. \cite{luo2018discriminability} method. The encoder has been trained with the COCO dataset. In fact, we utilized an image captioning system for this part. Some of the videos have been selected from the YouTube-8M dataset composed of almost 8 million videos totaling 500K hours of video \cite{abu2016youtube} and some from the COCO dataset \cite{lin2014microsoft}.
The captioning method has been trained and evaluated on the COCO dataset, which includes 113,287 images for training, 5,000
images for validation and another 5,000 held out for testing. Each image is associated with five human captions.
\begin{figure*}[htbp]
    \centering
    \includegraphics[width=\linewidth]{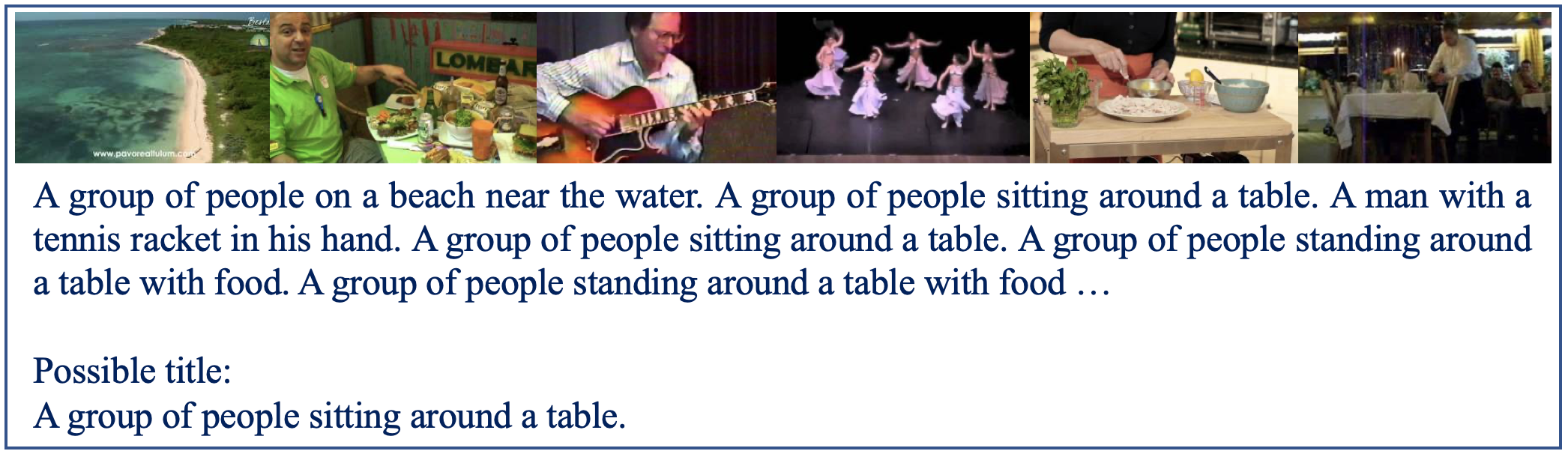}
    \centering
    \includegraphics[width=\linewidth]{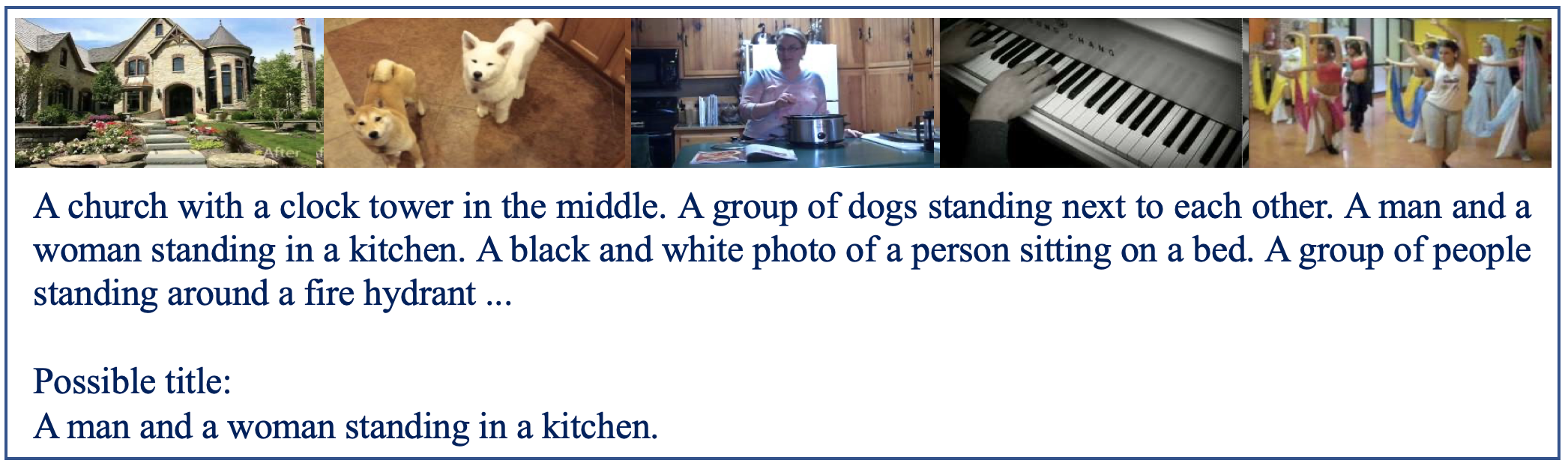}
    \centering
    \includegraphics[width=\linewidth]{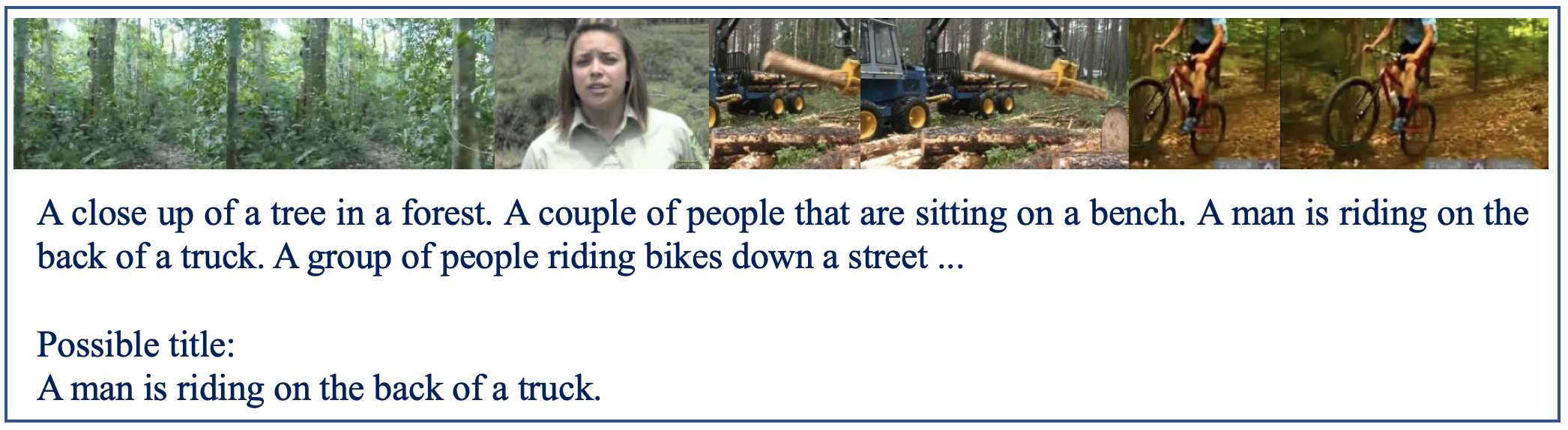}
    \centering
    \includegraphics[width=\linewidth]{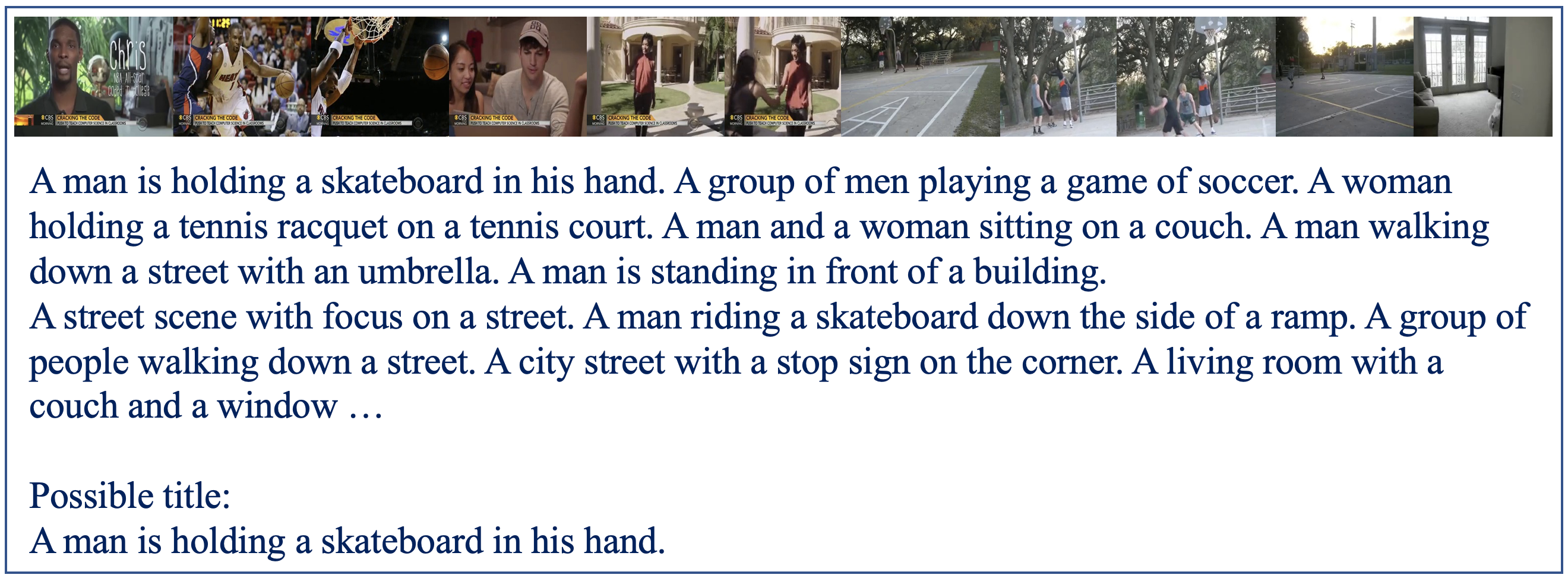}
    \caption{These are different videos of YouTube-8M and ActivityNet dataset that we captioned with \cite{luo2018discriminability} and made a document. Then, the possible title is generated with an extractive text summarization algorithm for each document.}
    \label{fig:txtSCST}
\end{figure*}

For the image encoder in the retrieval and FC captioning model, Resnet-101 is used. For each image, the global average pooling of the final convolutional layer output is used, results in a vector of dimension 2048.
The spatial features are extracted from the output of a Faster R-CNN with ResNet-101 \cite{amirian2018}, trained with the object and attribute annotations from Visual Genome \cite{krishna2017visual}. Both the FC features and Spatial features are pre-extracted, and no fine-tuning is applied to image encoders. For captioning models, the dimension of LSTM hidden state, image feature embedding, and word embedding are all set to 512. The retrieval model uses GRU-RNN to encode text. The word embedding has 300 dimensions and the GRU hidden state size and joint embedding size are 1024 \cite{luo2018discriminability}. The captions generated with this model describe valuable information about the frames. However, richer and more diverse sources of training signal may further improve the training of caption generators.


The Text Summarization method that has been used in the first experiment is extractive and single summarization. First, we read the generated document from the previous process. Then, we generate a Similarly Matrix across the sentences. We then rank the sentences in the similarity matrix. And at the end, we sort the rank and pick the top sentence. Figure \ref{fig:txtSCST} shows some experiments that have been done. The videos are selected from the YouTube-8M dataset \cite{abu2016youtube} and some from the COCO dataset \cite{lin2014microsoft}.
The reader can find all the guidance and code here\footnote{https://github.com/sohamirian/VideoTitle} for replicating the experiments for each part of the process.

\begin{figure}[htbp]
    \centering
    \includegraphics[width=\linewidth]{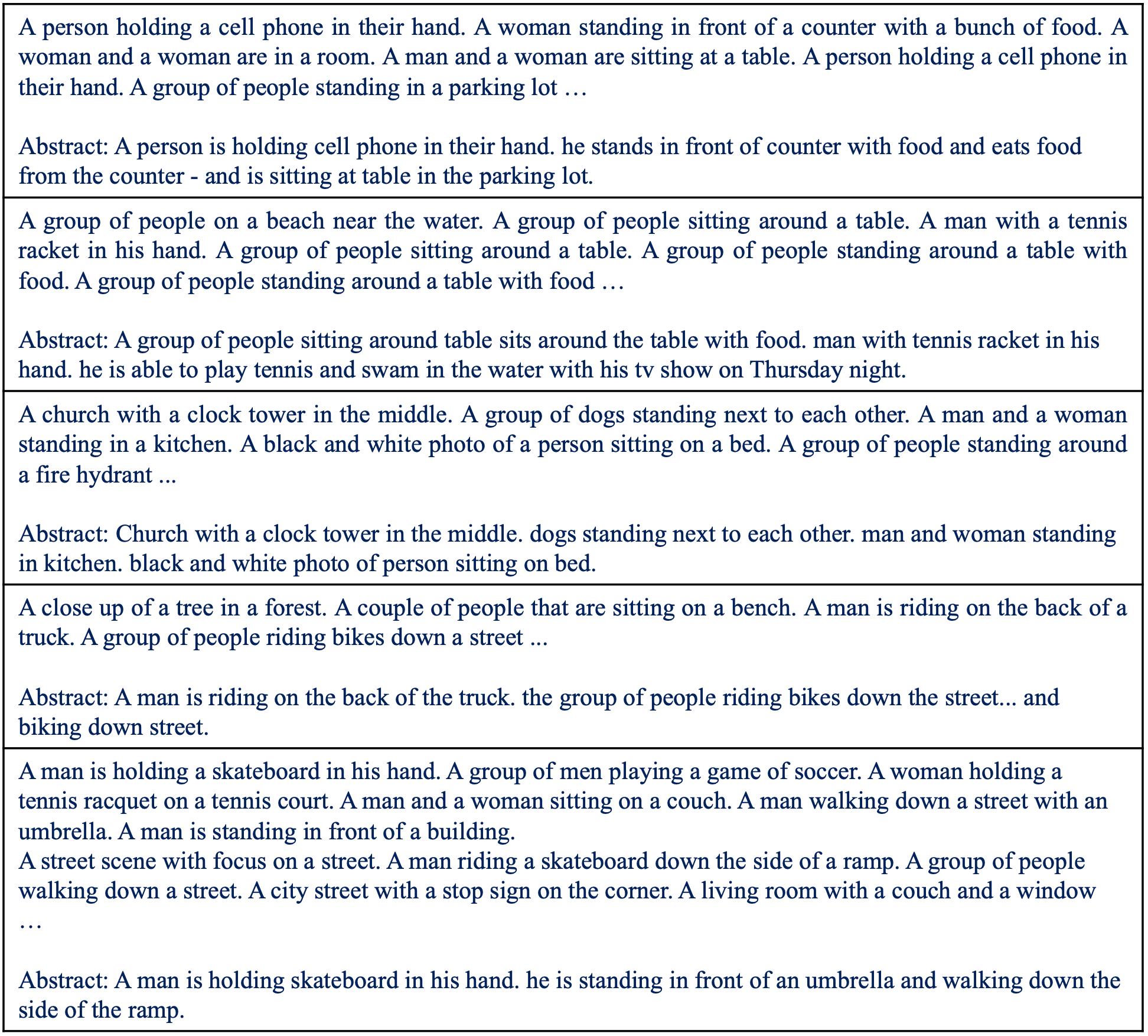}
    \caption{Here are the results with the abstractive text summarization method. Which generates a summary for each document.}
    \label{fig:txtSumAbs}
\end{figure}

In another experiment, we implement the abstractive text summarization method \cite{raffel2019exploring} to generate an abstract for the video clips, instead of assigning a title. Figure \ref{fig:txtSumAbs} shows the results by using Simple abstractive text summarization with pre-trained T5 (Text-To-Text Transfer Transformer) code \cite{absTxt}.

\section{Conclusion}
\label{sec:conclusion}

The purpose of this research is to propose an architecture that could generate an appropriate title and a concise abstract for a video by utilizing an image/video caption systems and text summarization methods to help in several domains such as search engines, supervision cameras, and the cinema industry. We utilized deep learning systems as captioning methods to generate a document describing a video. We then use extractive text summarization methods to assign a title and abstractive text summarization methods to create a concise abstract to the video. We explained the components of the proposed framework and conducted experiments using videos from different datasets. The results prove that the concept is valid. However, the results could become better by applying more improved frameworks of image/video captioning and text summarization methods. 

In our future work, we plan to explore more recent techniques of image/video captioning systems to generate a more natural story to describe the video clips. Therefore, the text summarization system could generate a better title using the extractive text summarization algorithms and a better abstract using the abstractive text summarization algorithms.

\section{Acknowledgement}
\label{sec:ack}
We gratefully acknowledge the support of NVIDIA Corporation with the donation of the Titan V GPU used for this research.

\bibliographystyle{spmpsci}

\end{document}